\documentclass[11pt]{article}
\usepackage[preprint]{acl}
\usepackage{times}
\usepackage{latexsym}
\usepackage[T1]{fontenc}
\usepackage[utf8]{inputenc}
\usepackage{microtype}
\IfFileExists{inconsolata.sty}{\usepackage{inconsolata}}{}
\usepackage{graphicx}
\usepackage{booktabs}
\usepackage{amsmath,amssymb}
\usepackage{tabularx}
\usepackage{listings}
\usepackage{tikz}
\usetikzlibrary{positioning,arrows.meta}
\title{MafiaScope: Inspecting Multi-Agent Reasoning\\
in Social Deduction Games}
\author{Ilia Karpov$^{1}$ \quad Maksim Grigorenko \quad Vadim Ayuyev$^{2}$ \\
  $^{1}$HSE University \quad $^{2}$Independent Researcher \\
  \texttt{karpovilia@gmail.com} \quad \texttt{mades.gri@gmail.com} \quad
  \texttt{vadim.ayuyev@gmail.com}}
\begin{document}
\maketitle
\begin{abstract}
An LLM agent's public behaviour reveals little about its social reasoning: an agent that votes correctly may be guessing, and its public claims may differ from privately elicited assessments. We present MafiaScope, an open testbed that turns the social deduction game Mafia into a measurement instrument for multi-agent reasoning. After each public utterance, living agents privately answer configurable structured questions on copies of their available conversations. Probing is non-invasive with respect to the acting context: neither the questions nor the answers enter the histories used to continue the game. Role assessments are scored against roles known to the engine; second-order predictions are compared with other agents' reports. An interactive visualizer renders the assessment trajectories: impersonate mode focuses on one agent's information and reports, panels chart timeline-aligned accuracy and calibration, and counterfactual replay branches from saved states. An offline heuristic links shifts in collective suspicion to episodes for inspection. Experiments span multiple models, languages, and probing configurations; the demonstration presents episode inspection, calibration, probe-repeatability and probe/no-probe checks, and utterance-replacement replay.\footnote{Code: \url{https://github.com/karpovilia/mafiascope}. Demo: \url{https://karpovilia.github.io/mafiascope/}. Video: \url{https://vimeo.com/1208920221}.}
\end{abstract}
\section{Introduction}
A lost game can reflect a wrong belief about another player, or a poor decision after correctly identifying that player's role. These cases call for different repairs. A researcher needs to reconstruct the information available to the agent, its role report, and the action that followed. Aligning these records across a long conversation is a practical difficulty in debugging agent teams \citep{epperson2025agdebugger}.\par
We built \textbf{MafiaScope} to make these distinctions observable during a game of Mafia. A hidden minority knows its teammates; the remaining players infer roles through discussion and vote to eliminate suspects. The engine knows every assigned role and controls which messages reach each player, providing a reference for evaluating their assessments.\par
The system combines three capabilities. \textbf{Isolated elicitation} asks structured questions on copies of agent contexts, keeping the questions and answers outside the histories used to act. \textbf{Reference-based scoring} checks role reports against the engine's hidden state and second-order predictions against other agents' reports. \textbf{Time-aligned inspection} connects these measurements to the messages available at each step and the actions that follow. Together, they support an investigation from a visible error to the reports and information surrounding it.\par
Changes in collective suspicion provide an entry point into long recordings. An offline selector links these shifts to the viewer, where impersonate mode opens the participating agents' perspectives. The walkthrough follows a shift toward Gray through the ensuing discussion and vote. Saved-state replay then supports comparison of alternative utterances.\par
The empirical material includes episode selection across 672 games, four case-study observations about role reports and voting, a 240-game probe/no-probe comparison, repeated probes at fixed states, and utterance-replacement replay. Six users also tested the viewer for finding self-contradictions and abrupt changes in assessments.\par
\begin{figure*}[t]
\centering
\includegraphics[width=.94\textwidth]{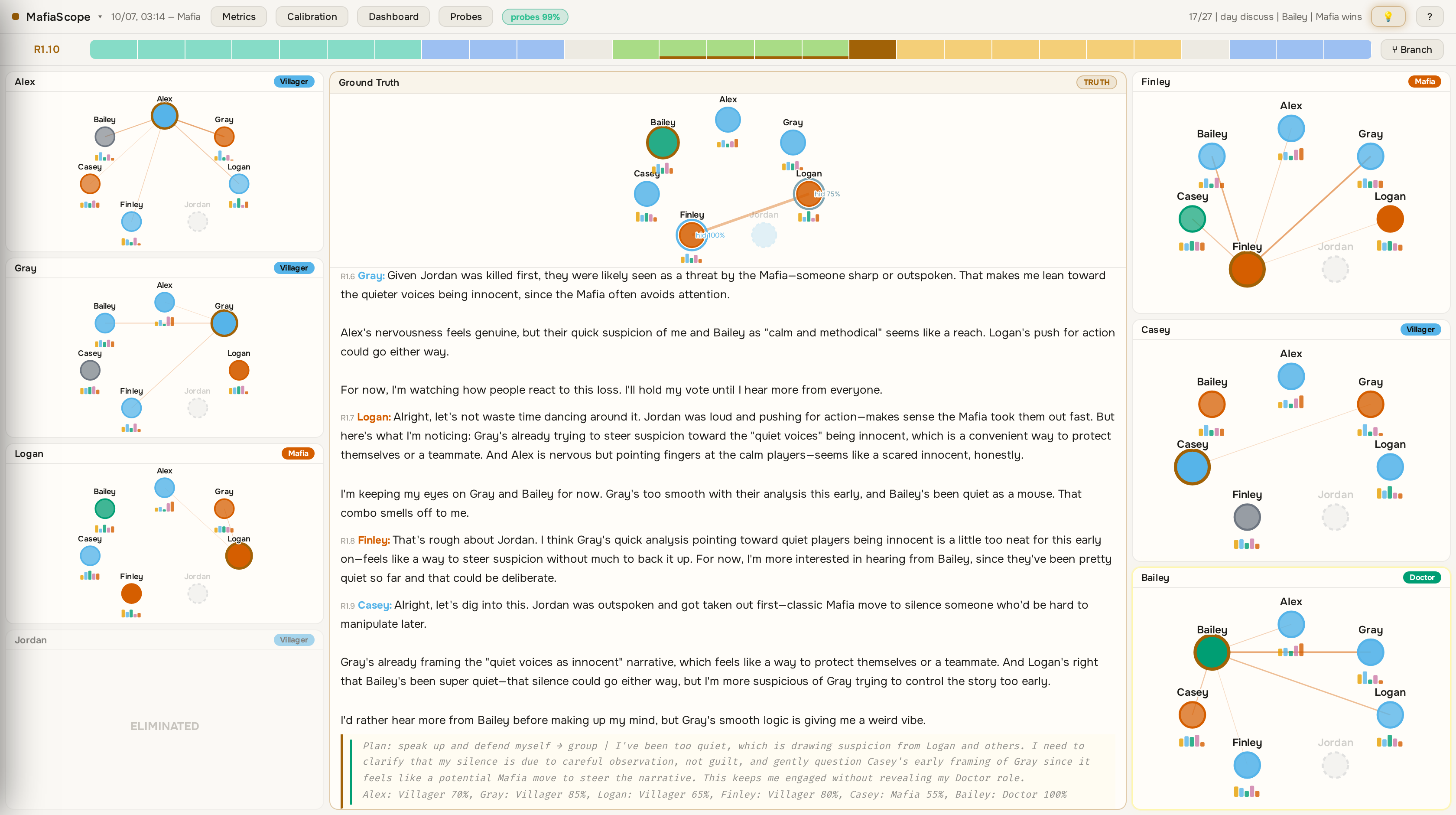}
\caption{The viewer aligns public messages, assigned roles, and agents' reports on a shared timeline.}
\label{fig:overview}
\end{figure*}
\section{Related Work}
Multi-agent inspection already has tools for navigating conversations and revisiting decisions. ChatArena supplies configurable language-game environments and a web interface \citep{wu2023chatarena}; Werewolf Arena exposes dialogue, private reasoning, bids, and votes \citep{bailis2024werewolfarena}. AGDebugger brings message editing and checkpoint resets into an interactive debugging workflow \citep{epperson2025agdebugger}. These capabilities establish much of the practical foundation for inspecting agent teams. Table~\ref{tab:comparison} locates MafiaScope's functions within this landscape.\par
The measurement problem is how to obtain comparable assessments while agents act with different information. Theory-of-Mind evaluation makes information asymmetry explicit in conversational tasks \citep{shinoda2025tomato,riemer2025position}; social-interaction benchmarks examine how agents pursue goals and infer intentions \citep{zhou2024sotopia,liu2024interintent}. Social deduction adds an engine-known hidden state. This permits role reports to be checked at the moment they are made, while second-order reports can be paired with the contemporaneous assessments of other players.\par
Work on Werewolf uses models of players' suspicions in planning \citep{zhang2025multimind}, follows honesty and deceptiveness judgments through dialogue \citep{agarwal2025wolf}, and examines belief--action discrepancies and responses to accusations \citep{gao2026auditing,yang2026accuser}. MafiaScope organizes these questions around a separate measurement channel: elicitation leaves the acting histories intact, scoring uses explicit references, and the viewer preserves the temporal connection between information, reports, and actions. This combination is the system's focus; replay supports the subsequent investigation.\par
\begin{table}[t]
\centering\small
\begin{tabularx}{\columnwidth}{@{}Xcccc@{}}
\toprule
Function & MS & CA & AG & WA \\
\midrule
Browse agent messages & $\checkmark$ & $\checkmark$ & $\checkmark$ & $\checkmark$ \\
Configure agent models & $\checkmark$ & $\checkmark$ & $\checkmark$ & $\checkmark$ \\
Probe copied contexts & $\checkmark$ & -- & -- & -- \\
Compare second-order reports & $\checkmark$ & -- & -- & -- \\
Plot role-confidence calibration & $\checkmark$ & -- & -- & -- \\
Edit a message and resume & $\checkmark^{*}$ & -- & $\checkmark$ & -- \\
\bottomrule
\end{tabularx}
\caption{Feature comparison: MafiaScope (MS), ChatArena (CA), AGDebugger (AG), Werewolf Arena (WA). Dashes mark undocumented functions; $^{*}$ denotes MafiaScope's local runner.}
\label{tab:comparison}
\end{table}
\section{Non-Invasive Probing}
\label{sec:system}
The game engine runs night actions, daytime discussion, and voting, with agent roles, personalities, language, and model backends configured independently. The demonstration uses seven-player games. Public events, probe responses, and optional snapshots are stored in three JSONL streams. The event key comprises the game, round, and public-message sequence; probe records additionally identify the agent and probe. Raw responses are retained alongside parsed values so that a displayed assessment can be checked against its source.\par
The probe engine builds a separate request from the agent's system prompt, its available game conversation, and a structured question. The question and answer are logged in the measurement channel and never appended to the acting agent's history. This context isolation defines \emph{non-invasive} in the observational configuration used here. The demonstration probes all living agents after each public utterance.\par
Probe chaining can pass earlier answers into later probe prompts, including across steps, while remaining within the measurement channel. The rendered prompts record this history. Role guesses are checked against assigned roles. Second-order predictions, such as who the agent thinks suspects it, are compared with other agents' reports. Table~\ref{tab:observations} in Appendix~\ref{app:triage} lists the observations.\par
Researchers configure the measurement battery in YAML, choosing each question, its trigger, and its token budget. The ranking probe used by the episode selector is declared as follows:
\begin{lstlisting}[basicstyle=\ttfamily\footnotesize,breaklines=true,frame=single]
- id: suspicion_ranking
  question: >
    Rank alive players ({players}) from
    most to least suspicious.
    Answer ONLY as JSON:
    [{{"player": "...", "score": N}}, ...]
\end{lstlisting}
The engine fills the player names at each step. Other supplied probes elicit role assignments and confidence, predictions about others' attitudes, planned actions, and personality assessments. A probe can run after each message, after a round, or on the responding agent's own turn. These triggers support both dense trajectories and measurements at selected decisions.\par
\section{The Visualizer}
\label{sec:visualizer}
\begin{figure*}[t]
\centering
\includegraphics[width=.94\textwidth]{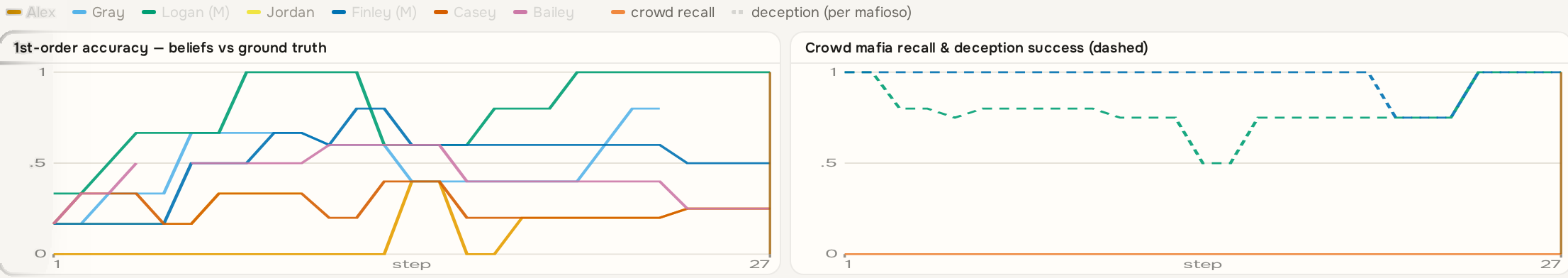}
\caption{Role accuracy (left) and majority-based Mafia detection (right). Dashed lines show the share of observers missing each Mafia player.}
\label{fig:metrics}
\end{figure*}

\begin{figure}[t]
\centering
\includegraphics[width=\columnwidth,trim=35bp 42bp 40bp 270bp,clip]{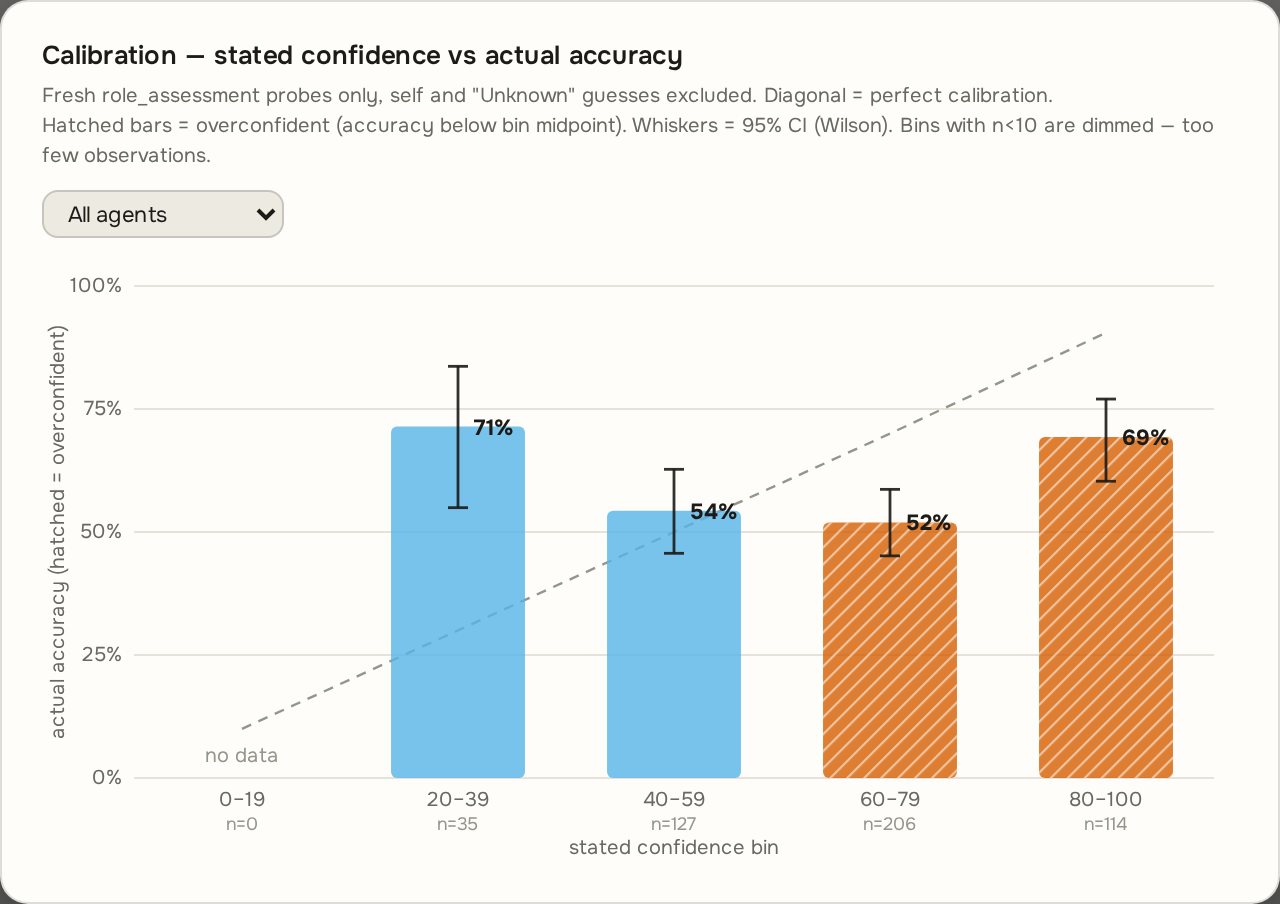}
\caption{Role accuracy by confidence in the walkthrough game. Dashed: perfect calibration; whiskers: Wilson 95\% intervals.}
\label{fig:calibration}
\end{figure}

Moving along the timeline updates the transcript and agent graphs (Figure~\ref{fig:overview}). Selecting an agent enters \emph{impersonate mode}, which focuses the display on its reports and available game information. Two ego views compare its predictions about other players' attitudes with their role reports about it. Hovering an edge reveals the explanation behind a reported relationship. Metric trajectories show how each agent's role accuracy changes through the conversation (Figure~\ref{fig:metrics}); clicking a point returns to that step. The calibration panel groups role guesses by stated confidence and compares them with assigned roles \citep{guo2017calibration,tian2023just}, excluding self-guesses and Unknown answers (Figure~\ref{fig:calibration}). Users can switch between individual agents and all agents, while the dashboard supports browsing across games.\par
Recorded games open in the browser without model access. The local replay server resumes play from a saved snapshot, while the experiment runner generates branches with replacement utterances. The viewer opens each branch for comparison with the original dialogue.\par
\subsection{Qualitative User Evaluation}
Six users tested the viewer by looking for model self-contradictions and abrupt changes in agents' assessments. Their qualitative feedback indicated that the system made it easier to find cases and problems to investigate for policy optimization. The evaluation collected no quantitative measures of search time or detection accuracy.\par
\section{From Collective Shifts to an Investigation}
\label{sec:workflow}
\subsection{Finding Candidate Episodes}
The offline selector compares consecutive discussion steps within a round, requiring the same living roster and a complete ranking with a unique maximum from each living agent at both steps. It uses event and probe logs; replay snapshots are optional. Appendix~\ref{app:triage} specifies parsing and eligibility rules.\par
The observer set includes all living agents, including Mafia, whose reports may be strategic. For an eligible set of agents $C$, let $z_{i,t}$ be agent $i$'s top suspect and let
\begin{equation}
q_t(j)=\frac{1}{|C|}\sum_{i\in C}\mathbf{1}[z_{i,t}=j].
\end{equation}
We record the number of agents that change their target, $K_t$, and the redistribution of support,
\begin{equation}
D_t=\frac{1}{2}\sum_j |q_t(j)-q_{t-1}(j)|.
\end{equation}
The script orders transitions by $D_t$, breaking ties with $K_t$, and records who switched and which targets led before and after. Ranking uses the reported top suspects; the walkthrough then follows the selected shifts through to voting.\par
Candidate records identify the game and message coordinates. In the demo, links open the preceding and following viewer states and each switching agent's perspective. Nearby candidates form a discussion window for inspection.\par
\subsection{Inspecting the Participating Agents}
\begin{figure*}[t]
\centering
\includegraphics[width=.94\textwidth]{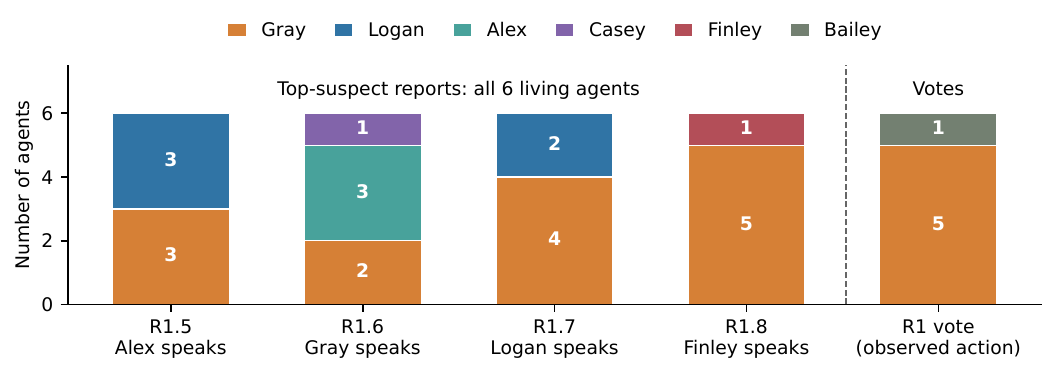}
\caption{Suspicion shifts toward Gray during discussion, followed by five votes for Gray. The final bar shows votes; earlier bars show top-suspect reports.}
\label{fig:shift}
\end{figure*}
The walkthrough opens one of the selected transitions, after Logan's R1.7 message, the seventh public-message step of round~1 (Figure~\ref{fig:shift}). Six of the original seven players remain alive after the first night. At the preceding step, three agents rank Alex first, two rank Gray first, and one ranks Casey first. After Logan speaks, four rank Gray first and two rank Logan first, giving $K=4$ and $D=4/6$. Finley's next message is followed by two further switches. Five agents now rank Gray first.\par
Opening Bailey's perspective shows one of these switches. Bailey's top suspect changes from Logan at R1.7 to Gray at R1.8. Its separate role assessment continues to call Gray a Villager, with confidence falling from 80 to 75. Later, Bailey votes for Gray. The ranking changes while the categorical role label stays fixed. Opening both explanations shows how each answer justifies its assessment.\par
The transcript shows Logan and Finley criticizing Gray's interpretation of the first night; later speakers repeat similar concerns. The ground-truth overlay identifies both speakers as Mafia. Their messages, Bailey's changing reports, and the subsequent responses are visible at the corresponding timeline steps.\par
\subsection{Checking Actions and Preparing Replay}
Advancing to the vote shows five votes for Gray and Gray's vote for Bailey. Gray is eliminated.\par
Gray's own vote shows a different discrepancy. Its preceding role report at R1.11 labels Logan as Mafia and Bailey as a Villager, but its R1.12 utterance ends with a vote for Bailey. The transcript places this disagreement beside Gray's explanation of its vote.\par
Gray's decision is a candidate for utterance-replacement replay from the preceding snapshot. The next section demonstrates this procedure with recorded voting decisions, including Dana's vote in another game. Opening a saved branch reveals the injected utterance and ensuing dialogue.\par
\section{The Bifurcation Experiment}
\label{sec:bifurcation}
The bifurcation experiment replaces a voter's full utterance at a saved decision point. For each decision, the protocol samples 500 candidate utterances at temperature 1.2. It selects 19 alternatives using vote-target coverage and farthest-point selection over multilingual E5 embeddings, and includes the original utterance as the twentieth variant. Each variant receives one continuation with probes disabled. Both wording and vote target can change. The viewer lists the variant text, parsed target, winner, and game length; clicking a row opens that branch's dialogue.\par
This replay analysis includes eight decisions selected using role reports and votes, and three selected at random, spanning five distinct parent games. The guided decisions are votes for non-Mafia players, grouped by whether the role report correctly names at least one Mafia player. Table~\ref{tab:bifurcation} reports the 220 continuations.\par
\begin{table}[t]
\centering\small
\begin{tabular}{@{}lrrr@{}}
\toprule
Selection / report & Points & Forks & V wins \\
\midrule
Guided: Mafia named & 3 & 60 & 3 \\
Guided: no Mafia named & 5 & 100 & 6 \\
Random selection & 3 & 60 & 17 \\
\bottomrule
\end{tabular}
\caption{Utterance-replacement outcomes. V wins: Villager-team wins across all variants, including the original utterance.}
\label{tab:bifurcation}
\end{table}
At Dana's round-2 decision, three of the 20 variants lead to Villager wins; the continuation with the original utterance ends in a Mafia win. Opening the branches reveals the subsequent messages and votes, giving the analyst concrete trajectories to compare with the original-utterance continuation.\par
\section{Case Studies}
\label{sec:evaluation}
Four observations from a 30-game Russian-language DeepSeek batch illustrate the measurements enabled by the system. F1, F2, and F4 concern non-Mafia agents; F3 pairs agents' predictions with contemporaneous reports. Appendix~\ref{app:cases} gives the measurement definitions and results for two additional recording batches.\par
\textbf{F1. Role reports change through play.} In the opening round, 91.6\% of role guesses are Unknown. Recall of actual Mafia players is 1.5\% in those reports and 62.5\% in the round-2 reports, which span 20 games. Recall counts correct Mafia identifications among assessments of living Mafia players, with Unknown counted as a miss. The timeline shows where these assessments become more definite and which messages precede the changes.\par
\textbf{F2. High confidence accompanies frequent errors.} Among 722 role assessments with confidence 80--100, mean stated confidence is 89.6\%, but accuracy is 61.4\%. These assessments span 22 games; 279 assign an incorrect role. The calibration panel shows the gap, while the transcript gives the explanations accompanying individual assessments.\par
\textbf{F3. Agents over-predict being suspected.} Across 5,301 same-step pairs, agents predict that another player suspects them 1,105 times; the paired role reports indicate suspicion 420 times. The ratio is 2.63 (95\% game-bootstrap CI [2.15, 3.25]). Each pair compares A's prediction that B suspects it with B's role report about A; a Mafia label at confidence at least 50 counts as suspicion. The two ego views make these disagreements inspectable for individual players.\par
\textbf{F4. Reported suspicion tracks subsequent votes.} The immediately preceding top-suspect report matches 176 of 204 votes (86.3\%), compared with 23.5\% for uniform selection among living non-self targets. The remaining 28 votes disagree with the unique highest-scoring target in the voter's preceding ranking. Cases such as Gray's contrary vote (Section~\ref{sec:workflow}) identify decisions where the report and action diverge, supplying candidates for replay.\par
\section{Instrument Validation}
\label{sec:validation}
\textbf{Context isolation.} Across 471 games with saved snapshots, the audit checks 113,794 distinct acting-context messages for private-probe and undeclared-feedback markers and finds none.\par
\textbf{Probed and unprobed play.} A randomized, interleaved comparison uses 240 games, 120 per arm. The arms share the seven-player setup, model backend, agent prompts, and sampling settings; the configurations differ only in probing. Mafia wins 84/120 games with probes and 89/120 without (Fisher's exact $p=.565$). The probed-minus-unprobed difference is $-4.2$ percentage points (95\% CI $[-15.3,+7.1]$). Both arms have a median of three rounds and 18 introduction or discussion messages, with Mann--Whitney $p=.929$ and $.604$, respectively.\par
\textbf{Probe repeatability.} Five repeated probes at each of nine saved states produce 42 disagreements across 90 answer pairs (46.7\%). Variation is concentrated in introductions and night actions. Appendix~\ref{app:checks} gives the phase-level counts and extraction rule.\par
\textbf{Recorded sample.} The observational corpus contains 672 completed games with discussion-time suspicion probes. The selector yields 1,422 complete adjacent-step comparisons across 361 games, including 727 with $D>0$. The five public demo games supply walkthrough examples within this larger sample. Appendix~\ref{app:triage} specifies the observation rules and recording settings. Selector tests check redistribution scores and target switches against fixed expected results.\par
\section{Availability}
The public repository provides the engine, probe definitions, viewer, and recorded examples under the project's MIT license. The static demo requires no credentials; generating games or branches requires a model backend. Locally, \texttt{main.py} records a game from a YAML configuration, and \texttt{serve\_viewer.py} prepares the viewer bundle and starts the browser interface with its replay server. Analysis scripts, inclusion decisions, and input hashes are included in the manuscript source package. The manifests identify the raw logs used by each analysis. Dense probing adds several model requests per agent and public step; users can reduce its frequency or select fewer probes.\par
\section*{Limitations}
Probe answers are elicited self-reports, sensitive to wording and potentially unfaithful to internal computation \citep{turpin2023language}. Probe chaining passes earlier answers into later questions, which can anchor subsequent reports.\par
Impersonate mode, log filtering, and suspicion thresholds help manage graph clutter, while the reading burden grows with the number of agents. Episode selection depends on complete rankings with unique maxima and favors sharp changes, so stable failures may remain unnoticed.\par
Replay changes both wording and vote target, with one continuation per variant and several decisions from the same parent game. Estimating intervention effects requires repeated controls. The corpus spans multiple models and languages with different prompts and sampling settings; comparative claims about reasoning quality require a matched protocol.\par
\section{Conclusion}
MafiaScope turns a social deduction game into a measurement instrument: an engine that indexes public events by game and step, a probe battery that questions agents without adding the questions or answers to their acting histories, a scorer that checks role reports against known roles, and a viewer that follows one agent's perspective and branches from saved states. The case studies show what outcome-only evaluation misses: confidently wrong role assessments, over-prediction of being suspected, and agreements and discrepancies between reports and votes. Replay supplies alternative continuations for examining selected decisions. Questions about role inference and questions about how reports translate into actions call for different follow-up experiments; telling them apart is what this workflow is for. Next steps are comparisons of probe configurations across matched model panels, repeated continuations per intervention, and a timed comparison of the viewer with raw-log inspection. Engine, probes, viewer, and recorded examples are MIT-licensed.\par
\clearpage
\section*{Ethics and Broader Impact}
The demonstrated games involve artificial agents in a fictional social-deduction setting. Tools for examining strategic communication have potential dual uses, including optimizing persuasive messages; the released records and replay branches support inspection of such interventions. AI tools assisted code inspection, analysis, and manuscript revision; the authors are responsible for the reported claims.\par
\bibliography{custom}
\appendix
\section{Reproducible Episode Selection}
\label{app:triage}
\begin{table}[t]
\centering\small
\begin{tabularx}{\columnwidth}{@{}lX@{}}
\toprule
Observation & Interpretation \\
\midrule
Role assessment & Guessed roles and stated confidence; compared with assigned roles. \\
Suspicion ranking & Scores over other players; the highest score defines a top suspect. \\
Social map & Who the agent says trusts or suspects it; compared with others' reports. \\
Planned action & A stated action and target, elicited on the agent's own turn. \\
Public vote & An action recorded by the engine. \\
\bottomrule
\end{tabularx}
\caption{Reports and actions recorded by MafiaScope. Each observation retains its agent, source step, and raw response.}
\label{tab:observations}
\end{table}
The read-only scripts in \texttt{reproducibility/} require no model calls. \texttt{audit\_shift\_coverage.py} inventories every local \texttt{logs/**/game.jsonl}; \texttt{audit\_game\_integrity.py} records whole-game admission decisions. \texttt{find\_shifts.py} retains detailed evidence for the five demo games and reproduces the walkthrough figure. Table~\ref{tab:coverage} groups the compared transitions by recording setting, counting unique game IDs.\par
\begin{lstlisting}[basicstyle=\ttfamily\footnotesize,breaklines=true,frame=single]
python reproducibility/audit_shift_coverage.py --repo /path/to/mafiascope
python reproducibility/audit_game_integrity.py --repo /path/to/mafiascope
python reproducibility/find_shifts.py --repo /path/to/mafiascope
python reproducibility/make_figures.py
\end{lstlisting}
\begin{table}[t]
\centering\small
\begin{tabularx}{\columnwidth}{@{}Xrr@{}}
\toprule
Recording setting & Games & Comparisons \\
\midrule
DeepSeek & 165 & 624 \\
GPT-4o-mini & 70 & 362 \\
Qwen3: ru & 18 & 25 \\
Qwen3: zh & 50 & 291 \\
Qwen3: de & 32 & 72 \\
Qwen3: es & 26 & 48 \\
\midrule
Total & 361 & 1,422 \\
\bottomrule
\end{tabularx}
\caption{Games and adjacent-step comparisons used by the selector, grouped by recording setting.}
\label{tab:coverage}
\end{table}
The observational sample consists of completed root games with valid game responses and discussion-time suspicion probes. The manifest records game IDs, backend settings, languages, and whole-game admission decisions. Alive rosters are reconstructed from setup and elimination events and checked against snapshots. The context-isolation check in Section~\ref{sec:validation} searches saved acting histories for the marker set listed by the audit script.\par
Each comparison uses adjacent \texttt{day\_discuss} events in the same round with an unchanged living roster. Every living observer supplies a current ranking at both events. The response is a JSON array, optionally inside a Markdown fence or a wrapper containing one list. Each living non-self target occurs once with a finite numeric score, and the highest score is unique. Probe timestamps lie after the associated public event and before the next public event. The selector reads the recorded scores directly at those steps.\par
The running example and the viewer figures use \texttt{36594b66}, selected for continuity with the public demonstration rather than maximum $D$. The Dana replay point uses \texttt{0da78714}. The running example's displayed window covers R1.5--R1.8, followed by the round-1 vote tally.\par
\section{Interpreting Instrument Checks}
\label{app:checks}
The A/B experiment uses a shared, interleaved pool of probed and unprobed games. Saved backend metadata identifies DeepSeek-v4-flash. \texttt{recheck\_ab\_lengths.py} recomputes game lengths from terminal round numbers and counts of \texttt{intro} and \texttt{day\_discuss} events, excluding voting utterances; its output records per-game values and input hashes.\par
The repeated-probe analysis uses nine states from five games. Each state supplies five answers and $\binom{5}{2}=10$ pairs. Table~\ref{tab:retestphases} groups comparisons by the recorded game phase. Introductions occur in round~0, the two voting states in round~1, and the night and discussion states in round~2. Pairs share answers within a state, and states can share a parent game. The per-state outputs are recorded in \texttt{prior\_evidence\_clean\_subset.json}.\par
\begin{table}[t]
\centering\small
\begin{tabular}{@{}lrr@{}}
\toprule
Phase & States & Disagreeing pairs \\
\midrule
Introduction & 4 & 30/40 \\
Night & 2 & 12/20 \\
Voting & 2 & 0/20 \\
Discussion & 1 & 0/10 \\
\midrule
Total & 9 & 42/90 \\
\bottomrule
\end{tabular}
\caption{Top-suspect disagreement across repeated probes at fixed states.}
\label{tab:retestphases}
\end{table}
The retest extraction reads JSON lists, keeps targets from the stored reference vector, and selects the maximum score with first-occurrence tie breaking. Each repeat supplies one top suspect for the within-state comparisons.\par
The calibration and timeline illustrations use the same recorded game as the walkthrough. Crowd recall counts living Mafia identified by a majority of living non-Mafia observers; each Mafia player's dashed series is the fraction of those observers not identifying it. The supplied input hashes identify the recording; API aliases such as \texttt{deepseek-chat} do not identify immutable model checkpoints.\par
\texttt{recheck\_bifurcation.py} reproduces Table~\ref{tab:bifurcation} from the whole-game admission decisions and raw continuation logs. It checks parent IDs, saved-state coordinates, unique variant indices, terminal events, and response integrity; the output records input hashes and per-decision outcomes. The guided and random subsets contain four and two parent games, respectively, with one shared parent. All 11 retained factual-utterance continuations end in Mafia wins. The guided groups contain three decisions with at least one correctly named Mafia player and five without one. The random subset contains two votes for actual Mafia and one for a non-Mafia player. The analysis subset and per-point outcomes are identified in the accompanying manifest.
\section{Case-Study Measurements}
\label{app:cases}
\texttt{case\_studies.py} reads public-step probes from the recorded batches. Role assessments use living non-Mafia observers and living non-self targets. F1 includes Unknown responses; F2 groups committed role guesses by stated confidence. The opening-round counts in the main batch are 2,315 Unknown guesses out of 2,526 assessments and 13 correctly identified Mafia roles out of 842 Mafia-target assessments. Round~2 contributes 125 correct Mafia identifications out of 200.\par
For F3, an agent's social-map prediction about another player is paired with that player's role report at the same game, round, public-message step, and phase. A Mafia guess at confidence at least 50 maps to ``suspects''; a Villager, Doctor, or Sheriff guess at that confidence maps to ``trusts''; other assessments map to ``neutral''. Pairs cover distinct living players other than Mafia teammates. The reported ratio divides predicted ``suspects'' labels by the corresponding labels in the paired role reports.\par
F4 takes the voter's ranking at the immediately preceding public step in the same round, with a complete living-target list and a unique maximum score. The engine's vote tally supplies the observed action. The baseline selects uniformly among living non-self targets at that vote.\par
\begin{table}[t]
\centering\small
\begin{tabularx}{\columnwidth}{@{}Xrrr@{}}
\toprule
Measurement & DS-RU & DS-mixed & GPT-mini \\
\midrule
Games & 30 & 10 & 74 \\
F1: round-0 Unknown (\%) & 91.6 & 75.1 & 97.6 \\
F1: round-2 Mafia recall (\%) & 62.5 & 65.4 & 30.1 \\
F2: high-confidence accuracy (\%) & 61.4 & 57.7 & 49.0 \\
F3: predicted/reported suspicion & 2.63 & 1.53 & 1.62 \\
F4: report--vote matches & 176/204 & 28/47 & 452/456 \\
\bottomrule
\end{tabularx}
\caption{Case-study observations by recording batch. DS denotes DeepSeek; GPT-mini denotes GPT-4o-mini.}
\label{tab:cases}
\end{table}
The manifest identifies the batches as \texttt{main30}, \texttt{corpus32}, and \texttt{gpt4omini100}, with the analyzed game IDs listed explicitly. Round-2 role reports span 20, 5, and 49 games, respectively. High-confidence assessments number 722, 743, and 8,714; the corresponding mean confidences are 89.6\%, 85.6\%, and 86.5\%. F3 uses 5,301, 7,634, and 57,981 paired reports. Intervals resample whole games 10,000 times. Per-game counts, input hashes, and all intervals are included in \texttt{case\_studies.json}.\par
\section{Replay Coordinates}
Utterance replacement starts from the snapshot preceding the utterance. The saved state contains the players' contexts and engine state at that point; each generated branch records the injected text and subsequent events.
\section{Relation to the Preprint}
The engine, probes, viewer, and replay mechanism were introduced in our preprint \citep{karpov2026mafiascope}. This demonstration adds collective-shift selection, a worked investigation, and reproducible measurement checks with explicit recording and observation criteria.
\end{document}